\documentclass{INTERSPEECH2023}
\usepackage{multirow}
\usepackage{cite}

\interspeechcameraready

\title{Transfer Learning from Pre-trained Language Models Improves \\End-to-End Speech Summarization}
\name{Kohei Matsuura, Takanori Ashihara, Takafumi Moriya, Tomohiro Tanaka,\\Takatomo Kano, Atsunori Ogawa, Marc Delcroix}
\address{
  NTT Corporation, Japan}
\email{kohei.matsuura@ntt.com}

\begin{document}

\maketitle
 
\begin{abstract}
End-to-end speech summarization (E2E SSum) directly summarizes input speech into easy-to-read short sentences with a single model. This approach is promising because it, in contrast to the conventional cascade approach, can utilize full acoustical information and mitigate to the propagation of transcription errors. However, due to the high cost of collecting speech-summary pairs, an E2E SSum model tends to suffer from training data scarcity and output unnatural sentences. To overcome this drawback, we propose for the first time to integrate a pre-trained language model (LM), which is highly capable of generating natural sentences, into the E2E SSum decoder via transfer learning. In addition, to reduce the gap between the independently pre-trained encoder and decoder, we also propose to transfer the baseline E2E SSum encoder instead of the commonly used automatic speech recognition encoder. Experimental results show that the proposed model outperforms baseline and data augmented models.
\end{abstract}
\noindent\textbf{Index Terms}: end-to-end speech summarization, transfer learning, pre-trained language model

\section{Introduction}

Speech summarization (SSum) technology is attracting increasing attention \cite{1,2,3} because its written style and short summaries are more user-friendly compared with word-by-word transcriptions. A typical way of realizing SSum is the cascade approach, where a text summarization (TSum) model summarizes input text transcribed by an automatic speech recognition (ASR) system. The cascade approach is successful thanks to a highlyaccurate ASR model \cite{4} and a TSum model pre-trained with a large amount of unpaired text data \cite{5}. However, this approach suffers from ASR error propagation \cite{6} and a lack of acoustical information that conveys the speaker’s attitude and emotion \cite{7}. To alleviate these limitations, we can exploit N-best ASR hypotheses to mitigate the effect of ASR errors \cite{8}, and add acoustic features to the input text \cite{9}.

As an alternative and more straight-forward approach, endto-end (E2E) SSum was proposed more recently. E2E SSum directly generates abstractive summaries from speech with a single model. Hence, it does not depend on erroneous transcriptions and can make full use of acoustical information. \cite{10} implements an E2E SSum model with restricted self-attention to deal with long speech input and reported that it outperforms the cascade model on the How2 dataset \cite{11}, which is often used for SSum experiments. In spite of the promising results, an E2E SSum model requires a large amount of costly speech-summary pairs for training, and tends to suffer from data scarcity and generate unnatural sentences. In \cite{12}, the authors tackle with this problem by using a text-to-speech (TTS)-based data augmentation approach \cite{13,14}. They converted an external TSum corpus into additional speech-summary pairs by the TTS technology and achieved a better performance than their baseline systems. However, the amount of training data may still not be enough for the E2E SSum model to generate natural sentences. 

\begin{figure}[t]
  \centering
  \includegraphics[width=\linewidth]{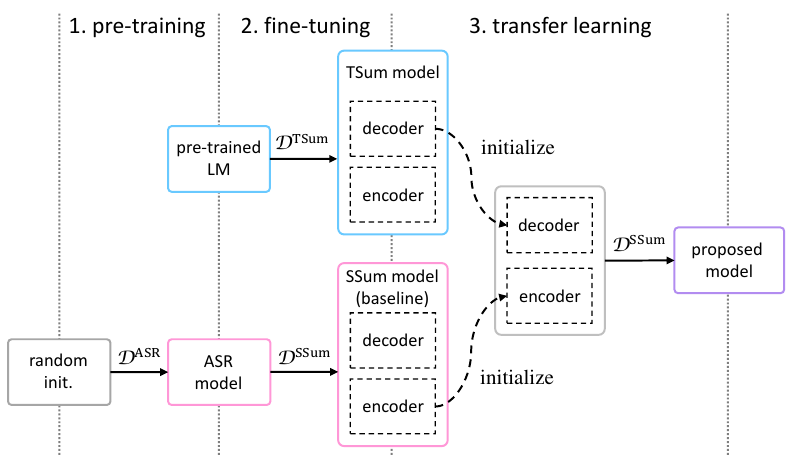}
  \caption{Our proposed method consists of three stages: pre-training, fine-tuning, and transfer learning. Arrows indicate which model is used for initialization. $\mathcal{D}^{\mathrm{ASR}}$, $\mathcal{D}^{\mathrm{TSum}}$, $\mathcal{D}^{\mathrm{SSum}}$ denote training data for ASR, TSum, and SSum, respectively.\vspace{-16pt}}
  \label{fig:speech_production}
\end{figure}

\begin{figure*}[t]
  \centering
  \includegraphics[width=\linewidth]{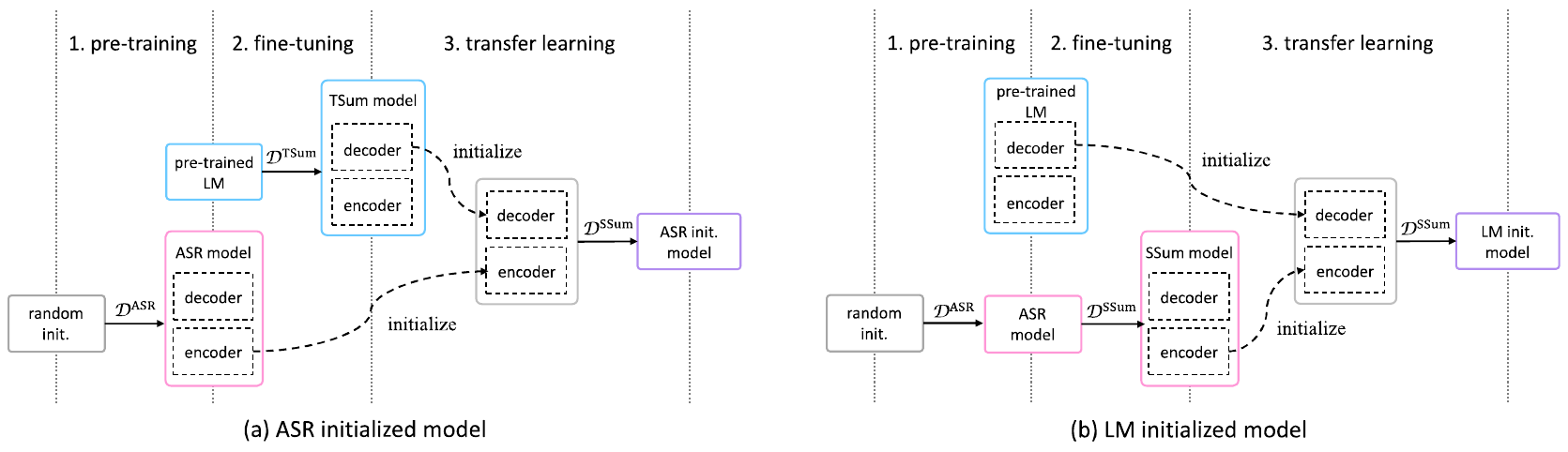}
  \caption{Two variants of our proposed model: (a) ASR initialized model and (b) LM initialized model.\vspace{-15pt}}
  \label{fig:speech_production}
\end{figure*}

Recently, large pre-trained language models (LMs) have been publicly and readily available \cite{5,15}. These models can be trained on an extremely large text corpus and have strong LM capabilities that allow them to generate very natural text. 
Encouraged by this, we propose for the first time to integrate a large pre-trained LM into an E2E SSum model via transfer learning. 
The procedure is depicted in Figure 1. 
First, we pre-train an ASR model and prepare a publicly available large pre-trained LM\footnote{We do not leverage self-supervised speech models for the encoder because the How2 dataset, which we used in the experiment, has only extracted acoustic features and does not contain raw waveforms. Thus, we apply ASR pre-training instead of speech self-supervised learning
in this work.}, both of which have an encoder-decoder architecture \cite{16}. Second, we fine-tune these models to obtain E2E SSum and TSum models. Finally, we transfer the SSum encoder and TSum decoder into a single model, which is further trained with a target SSum dataset (i.e., transfer learning) without freezing any parameters. Our proposed method for E2E SSum has two major differences from the typical transfer learning approach for other E2E speech processing tasks (e.g., E2E speech translation \cite{17,18,19,20,21}):
\begin{enumerate}
\item For the decoder initialization, we leverage a strong TSum model fine-tuned from a large pre-trained LM. Since the output sentences should be in written style and sophisticated in the SSum task, the capability of a pre-trained LM to generate natural sentences is expectedly essential.
\item For the encoder initialization, we transfer the baseline E2E SSum encoder instead of the commonly used ASR encoder. We assume that the encoder should be fine-tuned on a task similar to TSum in advance and more powerful to effectively collaborate with the strong TSum decoder.
\end{enumerate}

To evaluate the proposed method, we conducted several experiments on the How2 dataset. The proposed method significantly improved METEOR scores by 1.4 and 1.2 points compared with the baseline model and TTS-based data augmentation method, respectively. In addition, our proposed model generated more natural summaries than other E2E SSummodels as we expected. We further reveal that we can choose to reduce the training cost by omitting TSum fine-tuning with a slight performance degradation.

\vspace{-5pt}
\section{Related Work}

The transfer learning approach is widely studied for the E2E speech translation task, where a single model directly translates speech utterances into transcriptions in a target language \cite{22}. For instance, \cite{17} reports that a Long Short-Term Memorybased speech translation model gains a large improvement when it is initialized with the ASR encoder and machine translation decoder. Some studies re-think the speech translation model architecture to avoid wasting unused subnets such as the ASR decoder and the machine translation encoder \cite{18,19}. \cite{20} leverages an unsupervisedly pre-trained speech encoder (wav2vec 2.0 \cite{23}) and a text decoder (mBART \cite{24}) and achieves state-of-the-art speech translation performance. Inspired by these successes, we propose to apply the transfer learning approach to E2E SSum, which to the best of our knowledge has not been attempted before. 

In \cite{21}, they point out that ASR pre-training is not enough
for speech translation tasks, and they additionally pre-train the encoder with two tasks: frame-based masked LM and lexicon translation. Motivated by this study, we also propose to transfer an SSum encoder, which is expectedly stronger than an ASR encoder.

\vspace{-7.5pt}
\section{Method}

\subsection{End-to-End Speech Summarization (E2E SSum)}

In this paper, we adopt the Transformer-based encoder-decoder model to implement an E2E SSum model. The encoder $\mathrm{Enc}(\cdot)$ receives a sequence of acoustic features $x_{1:T}$ and outputs a sub-sampled sequence of hidden representations $\mathbf{h}^{\mathrm{Enc}}_{1:[\frac{T}{r}]}$ with a sub-sampling rate of $r$:
\begin{eqnarray}
\mathbf{h}^{\mathrm{Enc}}_{1:[\frac{T}{r}]} = \mathrm{Enc(\mathbf{x}, \mathbf{R}; \theta_{\mathrm{enc}})},
\end{eqnarray}
where $\theta_{\mathrm{enc}}$ is the parameters of the encoder. 
We use sub-sampling convolutional neural networks (CNNs) followed by Conformer blocks \cite{25} as $\mathrm{Enc}(\cdot)$. 
$\mathbf{R}$ denotes relative positional embeddings.

The decoder $\mathrm{Dec}(\cdot)$ receives $\mathbf{h}^{\mathrm{Enc}}_{1:[\frac{T}{r}]}$ and previous output tokens $\hat{y}_{1:l-1}$ to estimate the $l$-th token $\hat{y}_l$:
\begin{gather}
\mathbf{h}^{\mathrm{Dec}}_l = \mathrm{Dec}([\mathrm{Emb}(\hat{y}_i) + \mathrm{LPE(i)}]^{i-1}_{i=0}, \mathbf{h}^{\mathrm{Enc}}_{1:[\frac{T}{r}]}; \theta_{\mathrm{dec}}), \\
\hat{\mathbf{y}}^{\mathrm{dist}}_{l} = \mathrm{softmax}(\mathbf{h}^{\mathrm{Dec}}_l), \\
\hat{y}_l = \underset{y_l}{\operatorname{argmin}}(\hat{\mathbf{y}}^{\mathrm{dist}}_{l}),
\end{gather}
where $\mathrm{Dec}(\cdot)$ is composed of Transformer blocks with source-target attention, and $\theta_{\mathrm{dec}})$ is the parameters of the decoder. The token embedding layer $\mathrm{Emb}(\cdot)$ and the learnable position embedding layer $\mathrm{LPE}(\cdot)$ embed a token and its position, respectively. At the beginning of the inference, the decoder receives a special start-of-sentence token $\langle \mathrm{sos} \rangle$. It outputs tokens until a special end-of-sentence token $\langle \mathrm{eos} \rangle$ is emitted.

In the training stage, the parameters $\theta_{\mathrm{enc}})$ and $\theta_{\mathrm{dec}})$ are optimized by the stochastic gradient descent algorithm to minimize cross-entropy loss:
\begin{eqnarray}
\mathcal{L} = - \frac{1}{L}\sum^{L}_l\sum^{K}_k y^{\mathrm{LS}}_{l,k} \log \hat{y}^{\mathrm{dist}}_{l,k},
\end{eqnarray}
where $L$ and $K$ denote the number of tokens in a target summary and the vocabulary size, respectively. $y^{\mathrm{LS}}_{l,k}$ denotes the $k$-th element of the one-hot encoding of $y_l$ with label smoothing regularization \cite{26}. 
$\hat{y}^{\mathrm{dist}}_{l,k}$ denotes the $k$-th element of $\hat{\mathbf{y}}^{\mathrm{dist}}_{l}$.
Teacher forcing is used to stabilize model training, where the ground truth token $y_i$ is used in formula (2) instead of estimated $\hat{y}_i$.

\subsection{TTS-based Training Data Augmentation}

One straight-forward approach to address the training data scarcity problem is TTS-based data augmentation \cite{12}. We experiment with this approach as another baseline. First, we synthesize raw waveforms from source text of an external text summarization dataset using a TTS model. Then, we extract acoustic features from the waveforms and obtain pairs of the synthesized-speech and corresponding summaries. Finally, we fine-tune the baseline SSum model with both real and synthesized data. As a drawback, this method can leverage only external TSum datasets, not unpaired text nor large pre-trained LMs. Furthermore, the speech synthesis incurs extra computational
costs.

\begin{table*}[h]
\centering
\caption{Results of baseline models (C-1, B-1, and B-2) and our proposed models (P-1, P-2, and P-3). We also note initialization parameters, e.g., proposed model (P-1) was initialized with SSum encoder and TSum decoder.}
\begin{tabular}{c|ccc|ccc}
\hline
\multirow{2}{*}{ID} & \multirow{2}{*}{Model} & \multicolumn{2}{c|}{Initialization} & \multirow{2}{*}{ROUGE-1, 2, L} & \multirow{2}{*}{METEOR} & \multirow{2}{*}{BERTScore} \\
                    &                        & Encoder            & Decoder        &                                &                         &                            \\ \hline
C-1                 & cascade (ASR+TSum)     & \multicolumn{2}{c|}{-}              & 61.1, 43.3, 55.7               & 30.5                    & 92.88                      \\
B-1                 & baseline (E2E SSum)    & \multicolumn{2}{c|}{ASR}            & 64.9, 49.6, 60.8               & 33.0                    & 93.55                      \\
B-2                 & data augmentation      & \multicolumn{2}{c|}{SSum}           & 65.3, 50.7, 61.3               & 33.2                    & 93.62                      \\ \hline
P-1                 & proposed               & SSum (B-1)         & TSum           & \textbf{67.0}, 52.1, \textbf{63.2}               & \textbf{34.4 }                   & \textbf{93.98}                      \\
P-2                 & ASR initialized        & ASR                & TSum           & 64.0, 48.4, 59.9               & 32.5                    & 93.49                      \\
P-3                 & LM initialized         & SSum (B-1)         & LM             & \textbf{67.0}, \textbf{52.3}, \textbf{63.2}               & 34.2                    & 93.85                      \\ \hline
\end{tabular}
\end{table*}

\subsection{Transfer Learning for E2E SSum}
Our proposed method uses three datasets: speech-transcription pairs $\mathcal{D}^{\mathrm{ASR}}$ for ASR, speech-summary pairs $\mathcal{D}^{\mathrm{SSum}}$ for SSum, and transcription-summary pairs $\mathcal{D}^{\mathrm{TSum}}$ for TSum.

As depicted in Figure 1, the procedure of our proposed method consists of the following three stages:
\begin{enumerate}
    \item First, we obtain pre-trained ASR and LMs as follows,
    \begin{itemize}
        \item A randomly initialized encoder-decoder model is trained with $\mathcal{D}^{\mathrm{ASR}}$ to obtain an ASR model.
        \item Instead of pre-training a large LM by ourselves, we leverage publicly available models. In this paper, we adopt the BART base model\footnote{https://huggingface.co/facebook/bart-base} \cite{5}.
    \end{itemize}
    \item Second, we fine-tune the pre-trained models to obtain the SSum and TSum models as follows,
    \begin{itemize}
        \item The ASR model is fine-tuned with $\mathcal{D}^{\mathrm{SSum}}$ to obtain an SSum model. This is the baseline E2E SSum model.
        \item The LM is fine-tuned with $\mathcal{D}^{\mathrm{TSum}}$ to obtain a TSum model.
    \end{itemize}
    \item Finally, we transfer the SSum encoder and the TSum decoder into an encoder-decoder model. Then, it is fine-tuned with $\mathcal{D}^{\mathrm{SSum}}$ to obtain the proposed model.
\end{enumerate}

The proposed model expectedly inherits the strong capability to encode speech from the SSum encoder and to generate summary sentences from the TSum decoder on the basis of a large pre-trained LM. In addition, this method does not require additional costs to synthesize and store the augmented data. 

In addition to the proposed model, we additionally examine two variants: the \textit{ASR initialized model} and the \textit{LM initialized model}. The ASR initialized model is initialized with the ASR encoder instead of the SSum encoder in the transfer learning stage as illustrated in Figure 2-(a). We can examine whether the SSum fine-tuning is necessary for our proposed method with this model. The LM initialized model is initialized with the LM decoder instead of the TSum decoder as illustrated in Figure 2(b). We expect that the LM model is already capable enough to generate target summaries, and we can further save the computational cost required to train the TSum model.

\section{Experiments}

\subsection{Datasets}

In all experiments, we used the How2 dataset \cite{11}, which is composed of short instructional speech and corresponding transcriptions and summaries. The data $\mathcal{D}^{\mathrm{ASR}}$, $\mathcal{D}^{\mathrm{SSum}}$, and $\mathcal{D}^{\mathrm{TSum}}$ in Section 3.3 were the speech-transcription, speech-summary, and transcription-summary pairs of this dataset, respectively.
The training, validation, and evaluation sets contained 68,336, 1,600, and 2,127 samples (i.e., triplets of speech, transcription, and summary), respectively. The input speech was truncated up to 100 seconds and composed of 40-dimensional log Mel-filterbank energies with 3-dimensional pitch features. The total input speech length was 1,700 hours, and the average length was 84.9 seconds. We also examined whether our proposed method works in several lower resource settings. We randomly sub-sampled 100-, 500-, and 1,000-hour subsets of the training set, which contained 4,243, 21,311, and 42,549 samples, respectively. We used these subsets in the fine-tuning and transfer learning stages. In the ASR pre-training stage, the full dataset was used because speech-transcription pairs are generally easy to collect. For the data augmentation explained in Section 3.2, we used the Gigaword corpus \cite{27,28}, which is composed of 3.8M pairs of first sentences and titles of news articles. The TTS model synthesized 12,000 hours of speech from this dataset.

\begin{table*}[h]
\caption{Summaries and METEOR scores obtained by Model B-1, B-2, and P-1 for sample ``vNGAOtftE'' in evaluation set. Model B-1 and B-2 generated awkward expressions as highlighted in bold, while Model P-1 generated more natural summary.}
\begin{tabular}{c|c|l}
\hline
ID   & METEOR & \multicolumn{1}{c}{Summary}  \\ \hline
Ref. & -      & \begin{tabular}{l}\scriptsize{When buying a computer, consider what the computer will be used for and modify the computer to fit specific needs. Get more memory}\\ \scriptsize{for graphic art and a faster processor for gaming on a computer with information from a computer and technology specialist in this free}\\ \scriptsize{video on computers.}   \end{tabular}                                                    \\ \hline
B-1  & 20.1   & \begin{tabular}{l}\scriptsize{\textbf{Buying a computer is a complicated process, but sometimes it’s a complicated process}, such as writing letters, surfing, internet,} \\ \scriptsize{checking email and spread sheets. Purchase a computer with tips from a computer and technology specialist in this free video} \\ \scriptsize{on computers.}  \end{tabular}                                                                      \\ \hline
B-2  & 22.0   & \begin{tabular}{l}\scriptsize{When buying a computer, it’s important to use a computer for basic every day office tasks like writing letters, \textbf{surfing the internet,}} \\ \scriptsize{\textbf{surfing the internet, surfing or even mass.} Find out how to buy a computer and how much space to put on a computer with information} \\ \scriptsize{from a computer and technology specialist in this free video on computers.} \end{tabular}\\ \hline
P-1  & 26.8   & \begin{tabular}{l}\scriptsize{When buying a computer, make sure to have a computer with a graphics card, a quality hard drive and lots of storage space for all the} \\ \scriptsize{equipment and supplies a computer has. Buy a computer with a demonstration from a computer and technology specialist in this free} \\ \scriptsize{video on computers.}\end{tabular} \\\hline                                                      
\end{tabular}\vspace{-10pt}  
\end{table*}

\subsection{Training and Evaluation}

We adopted two encoder-decoder models in these experiments. For the ASR and E2E SSum models, the encoder was composed of 4-layer CNNs with a sub-sampling rate of 4 and 12-layer Conformer blocks with a model dimension of 768. The Conformer blocks had 8 attention heads, 2048-dimensional feed-forward (FF) layers, and a kernel size of 31. The batch normalization layers in the convolution modules were replaced by layer normalization to stably deal with small batch sizes. For the TSum model, the encoder was composed of 6-layer Transformer blocks with 12 attention heads and 3072-dimensional FF layers. The decoder structure was completely common for both E2E SSum and TSum. It was composed of 6-layer Transformer blocks with 12 attention heads and 3072-dimensional FF layers. The SSum (ASR) and TSum models have 280M and 140M parameters, respectively.

In all the stages in Section 3.3, we used the Adam optimizer \cite{29}. In the ASR pre-training stage, we used the Xavier initialization \cite{30} and Noam scheduler \cite{16} with a maximum learning rate of 2x10$^{-3}$, 40,000 warm-up steps, a weight decay rate of 10$^{-6}$, and a batch size of 512. The word error rate of the ASR model was 9.8\% with a beam size of 16 on the evaluation set. In the SSum fine-tuning and transfer learning stages, we halved the learning rate starting at 1x10$^{-4}$ if the validation loss was not improved. The batch size was 30. We applied SpecAugment \cite{31} to input features in these three stages. In the TSum fine-tuning stage, we linearly decayed the learning rate starting at 5x10$^{-5}$ for 20 epochs with a batch size of 8. We adopted the vocabulary of the public BART model introduced in Section 3.3 (i.e., Byte-Pair Encoding \cite{32} with a size of 50,265) in all experiments for consistency between pre-training
and fine-tuning.

For the data augmentation in Section 3.2, we followed the setting described in \cite{12} except for the vocabulary. We used VITS \cite{33} trained with the LJSpeech dataset \cite{34} for the TTS model. During fine-tuning, the learning rates started at 1x10$^{-4}$ and 1x10$^{-3}$ on the encoder and decoder, respectively. The maximum total length of input sequences in one batch was set to 300k. Each batch contained only real or artificial samples. The other optimization settings were the same as in the above SSum fine-tuning stage. We also implemented the cascade model as another conventional baseline. The ASR and TSum models were identical to those above mentioned.

For evaluation, we used the checkpoint that achieved the best validation accuracies. The beam width was 8, the length penalty was 0.3, and early end detection \cite{35} was applied. We chose ROUGE \cite{36}, METEOR \cite{37}, and BERTScore \cite{38} scores for the objective metrics, which are commonly used in TSum studies. We used ESPnet2 \cite{39} for all of the implementation, training, and evaluation.

\subsection{Results}

In Table 1, we show the ROUGE, METEOR, and BERTScore scores of the baseline models (C-1, B-1, and B-2) and proposed models (P-1, P-2, and P-3) on the How2 evaluation set. All the E2E models outperformed the cascade model (C-1). This result implies that it is important for SSum to leverage nonverbal information and to not depend on erroneous transcriptions. The augmented data model (B-2) improved the summarization accuracies even though the domain of the Gigaword dataset was fairly different from the target How2 dataset as reported in \cite{12}.

Our main proposed model (P-1) further improved the ROUGE-L, METEOR, and BERTScore scores by 2.4, 1.4, and 0.33 points from the E2E baseline model (B-1), resulting in the best performance in our experimental setting. When the ASR encoder was used for the initialization, the performance of our proposed method did not improve as seen in the results of the ASR initialized model (P-2), unlike the previous studies in the speech translation field \cite{17}. This result indicates that pre-training the encoder with the difficult SSum task was essential, and the ASR pre-training was not enough, unlike the previous studies in Section 2. In contrast, the results of the LM initialized model (P-3) show that our method worked well without the TSum fine-tuning stage. Additionally, both Model P-1 and P-3 required almost the same number of fine-tuning steps to converge in the transfer learning stage. Hence, we can reduce the training time with a small degradation of performance by using the pre-trained LM (i.e., BART) decoder directly.

We show examples of summaries and their METEOR scores obtained by Model B-1, B-2, and P-1 in Table 2. Model B-1 and B-2 output awkward sentences as highlighted in bold, but Model P-1 generated more natural sentences. This result suggests that the E2E SSum model should be trained with a larger amount of text than the external TSum dataset, while transferring the pre-trained LM was effective to improve the quality of summaries.

Finally, we show the METEOR scores obtained by Model P-1, B-1, and B-2 with various amounts of training data in Figure 3. The proposed model P-1 performed best with training data of 500 hours (21k samples) or more. When the amount of training data was only 100 hours (4.2k samples), the performance of the proposed model did not improve probably because it was difficult to integrate two independently pre-trained modules with the very small amount of training data.

\begin{figure}[t]
  \centering
  \includegraphics[width=\linewidth]{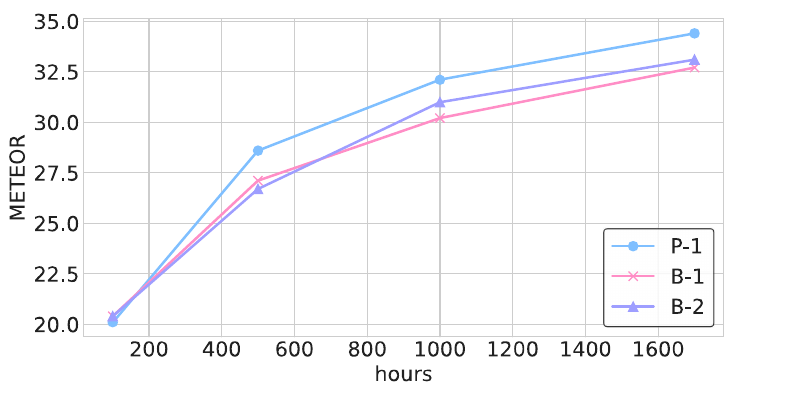}
  \caption{METEOR scores obtained by Model P-1, B-1, and B-2 with various amounts of training data.\vspace{-10pt}}
  \label{fig:speech_production}
\end{figure}

\section{Conclusion}

In this paper, we proposed to leverage a large pre-trained LM for E2E SSum via transfer learning. The proposed model achieved the best performance and generated the most natural sentences in our experiments. We found that it was essential to transfer the E2E SSum encoder instead of the ASR encoder. The results also showed that we can save on training time with little performance degradation when skipping the TSum fine-tuning stage. In addition, the proposed models achieved the highest performance with training data of 500 hours or more. Future research includes collecting a new speech summarization dataset with raw waveforms to examine the effect of integrating speech pre-trained models for the encoder.

\bibliographystyle{IEEEtran}
\bibliography{mybib}

\begin{thebibliography}{10}
\providecommand{\url}[1]{#1}
\csname url@samestyle\endcsname
\providecommand{\newblock}{\relax}
\providecommand{\bibinfo}[2]{#2}
\providecommand{\BIBentrySTDinterwordspacing}{\spaceskip=0pt\relax}
\providecommand{\BIBentryALTinterwordstretchfactor}{4}
\providecommand{\BIBentryALTinterwordspacing}{\spaceskip=\fontdimen2\font plus
\BIBentryALTinterwordstretchfactor\fontdimen3\font minus
  \fontdimen4\font\relax}
\providecommand{\BIBforeignlanguage}[2]{{%
\expandafter\ifx\csname l@#1\endcsname\relax
\typeout{** WARNING: IEEEtran.bst: No hyphenation pattern has been}%
\typeout{** loaded for the language `#1'. Using the pattern for}%
\typeout{** the default language instead.}%
\else
\language=\csname l@#1\endcsname
\fi
#2}}
\providecommand{\BIBdecl}{\relax}
\BIBdecl

\bibitem{1}
G.~Murray, G.~Carenini, and R.~Ng, ``Generating and validating abstracts of
  meeting conversations: a user study,'' in \emph{Proc. of ACL}, 2010.

\bibitem{2}
R.~Sharma, M.~Mahrishi, S.~Morwal, and G.~Sharma, ``Index point detection for
  text summarization using cosine similarity in educational videos,'' \emph{IOP
  Conference Series: Materials Science and Engineering}, 2021.

\bibitem{3}
G.~Finley, E.~Edwards, A.~Robinson, M.~Brenndoerfer, N.~Sadoughi, J.~Fone,
  N.~Axtmann, M.~Miller, and D.~Suendermann-Oeft, ``An automated medical scribe
  for documenting clinical encounters,'' in \emph{Proc. of NAACL}, 2018, pp.
  11--15.

\bibitem{4}
S.~Karita, N.~Chen, T.~Hayashi, T.~Hori, H.~Inaguma, Z.~Jiang, M.~Someki,
  N.~E.~Y. Soplin, R.~Yamamoto, X.~Wang, S.~Watanabe, T.~Yoshimura, and
  W.~Zhang, ``A comparative study on transformer vs rnn in speech
  applications,'' in \emph{ASRU}, 2019, pp. 449--456.

\bibitem{5}
M.~Lewis, Y.~Liu, N.~Goyal, M.~Ghazvininejad, A.~Mohamed, O.~Levy, V.~Stoyanov,
  and L.~Zettlemoyer, ``{BART}: Denoising sequence-to-sequence pre-training for
  natural language generation, translation, and comprehension,'' in \emph{Proc.
  of ACL}, 2020, pp. 7871--7880.

\bibitem{6}
N.~Ruiz and M.~Federico, ``Assessing the impact of speech recognition errors on
  machine translation quality,'' in \emph{Proc. of AMTA}, 2014, pp. 261--274.

\bibitem{7}
D.~Jouvet, ``Speech processing and prosody,'' in \emph{Proc. of TSD}, 2019, pp.
  3--15.

\bibitem{8}
T.~Kano, A.~Ogawa, M.~Delcroix, and S.~Watanabe, ``Attention-based
  multi-hypothesis fusion for speech summarization,'' in \emph{Proc. of ASRU},
  2021, pp. 487--494.

\bibitem{9}
T.-E. Liu, S.-H. Liu, and B.~Chen, ``A hierarchical neural summarization
  framework for spoken documents,'' in \emph{Proc. of ICASSP}, 2019, pp.
  7185--7189.

\bibitem{10}
R.~Sharma, S.~Palaskar, A.~W. Black, and F.~Metze, ``End-to-end speech
  summarization using restricted self-attention,'' in \emph{Proc. of ICASSP},
  2022, pp. 8072--8076.

\bibitem{11}
R.~Sanabria, O.~Caglayan, S.~Palaskar, D.~Elliott, L.~Barrault, L.~Specia, and
  F.~Metze, ``How2: A large-scale dataset for multimodal language
  understanding,'' in \emph{Proc. of NeurIPS Workshop ViGIL}, 2018.

\bibitem{12}
K.~Matsuura, T.~Ashihara, T.~Moriya, T.~Tanaka, A.~Ogawa, M.~Delcroix, and
  R.~Masumura, ``Leveraging large text corpora for end-to-end speech
  summarization,'' in \emph{Proc. of ICASSP}, 2023.

\bibitem{13}
T.~Kano, S.~Sakti, and S.~Nakamura, ``Structured-based curriculum learning for
  end-to-end english-japanese speech translation,'' in \emph{Proc. of
  INTERSPEECH}, 2017, pp. 2630--2634.

\bibitem{14}
M.~Mimura, S.~Ueno, H.~Inaguma, S.~Sakai, and T.~Kawahara, ``Leveraging
  sequence-to-sequence speech synthesis for enhancing acoustic-to-word speech
  recognition,'' in \emph{Proc. of SLT}, 2018, pp. 477--484.

\bibitem{15}
C.~Raffel, N.~Shazeer, A.~Roberts, K.~Lee, S.~Narang, M.~Matena, Y.~Zhou,
  W.~Li, and P.~J. Liu, ``Exploring the limits of transfer learning with a
  unified text-to-text transformer,'' \emph{Journal of Machine Learning
  Research}, 2020.

\bibitem{16}
A.~Vaswani, N.~Shazeer, N.~Parmar, J.~Uszkoreit, L.~Jones, A.~N. Gomez, L.~u.
  Kaiser, and I.~Polosukhin, ``Attention is all you need,'' in \emph{NeurIPS},
  2017.

\bibitem{17}
A.~B\'{e}rard, L.~Besacier, A.~C. Kocabiyikoglu, and O.~Pietquin, ``End-to-end
  automatic speech translation of audiobooks,'' in \emph{Proc. of ICASSP},
  2018, pp. 6224--6228.

\bibitem{18}
C.~Wang, Y.~Wu, S.~Liu, Z.~Yang, and M.~Zhou, ``Bridging the gap between
  pre-training and fine-tuning for end-to-end speech translation,'' in
  \emph{Proc. of AAAI}, vol.~34, 2020, pp. 9161--9168.

\bibitem{19}
Q.~Dong, M.~Wang, H.~Zhou, S.~Xu, B.~Xu, and L.~Li, ``Consecutive decoding for
  speech-to-text translation,'' in \emph{Proc. of AAAI}, May 2021, pp.
  12\,738--12\,748.

\bibitem{20}
X.~Li, C.~Wang, Y.~Tang, C.~Tran, Y.~Tang, J.~Pino, A.~Baevski, A.~Conneau, and
  M.~Auli, ``Multilingual speech translation from efficient finetuning of
  pretrained models,'' in \emph{Proc. of ACL}, 2021, pp. 827--838.

\bibitem{21}
C.~Wang, Y.~Wu, S.~Liu, M.~Zhou, and Z.~Yang, ``Curriculum pre-training for
  end-to-end speech translation,'' in \emph{Proc. of ACL}, 2020, pp.
  3728--3738.

\bibitem{22}
A.~B{\'e}rard, O.~Pietquin, L.~Besacier, and C.~Servan, ``Listen and translate:
  A proof of concept for end-to-end speech-to-text translation,'' in
  \emph{Proc. of NIPS Workshop on End-to-End Learning for Speech and Audio
  Processing}, 2016.

\bibitem{23}
A.~Baevski, Y.~Zhou, A.~Mohamed, and M.~Auli, ``wav2vec 2.0: A framework for
  self-supervised learning of speech representations,'' in \emph{Proc. of
  NeurIPS}, 2020, pp. 12\,449--12\,460.

\bibitem{24}
Y.~Liu, J.~Gu, N.~Goyal, X.~Li, S.~Edunov, M.~Ghazvininejad, M.~Lewis, and
  L.~Zettlemoyer, ``Multilingual denoising pre-training for neural machine
  translation,'' \emph{Trans. of ACL}, pp. 726--742, 2020.

\bibitem{25}
``Conformer: Convolution-augmented transformer for speech recognition,'' in
  \emph{Proc. of INTERSPEECH}, A.~Gulati, C.-C. Chiu, J.~Qin, J.~Yu, N.~Parmar,
  R.~Pang, S.~Wang, W.~Han, Y.~Wu, Y.~Zhang, and Z.~Zhang, Eds., 2020, pp.
  5036--5040.

\bibitem{26}
C.~Szegedy, V.~Vanhoucke, S.~Ioffe, J.~Shlens, and Z.~Wojna, ``Rethinking the
  inception architecture for computer vision,'' in \emph{Proc. of CVPR}, 2016,
  pp. 2818--2826.

\bibitem{27}
D.~Graff and C.~Cieri, ``English gigaword,'' \emph{Linguistic Data Consortium,
  Philadelphia}, 2003.

\bibitem{28}
A.~M. Rush, S.~Chopra, and J.~Weston, ``A neural attention model for
  abstractive sentence summarization,'' in \emph{Proc. of EMNLP}, 2015, pp.
  379--389.

\bibitem{29}
D.~P. Kingma and J.~Ba, ``Adam: A method for stochastic optimization.'' in
  \emph{Proc. of ICLR}, 2015.

\bibitem{30}
X.~Glorot and Y.~Bengio, ``Understanding the difficulty of training deep
  feedforward neural networks,'' in \emph{Proc. of AISTATS}, 2010, pp.
  249--256.

\bibitem{31}
D.~S. Park, W.~Chan, Y.~Zhang, C.-C. Chiu, B.~Zoph, E.~D. Cubuk, and Q.~V. Le,
  ``{SpecAugment}: A simple data augmentation method for automatic speech
  recognition,'' in \emph{Proc. of INTERSPEECH}, 2019, pp. 2613--2617.

\bibitem{32}
R.~Sennrich, B.~Haddow, and A.~Birch, ``Neural machine translation of rare
  words with subword units,'' in \emph{Proc. of ACL}, 2016, pp. 1715--1725.

\bibitem{33}
J.~Kim, J.~Kong, and J.~Son, ``Conditional variational autoencoder with
  adversarial learning,'' in \emph{Proc. of ICML}, 2021, pp. 5530--5540.

\bibitem{34}
K.~Ito and L.~Johnson, ``The lj speech dataset,''
  \url{https://keithito.com/LJ-Speech-Dataset/}, 2017.

\bibitem{35}
H.~Miao, G.~Cheng, P.~Zhang, and Y.~Yan, ``Online hybrid ctc/attention
  end-to-end automatic speech recognition architecture,'' \emph{IEEE/ACM
  Transactions on Audio, Speech, and Language Processing}, pp. 1452--1465,
  2020.

\bibitem{36}
C.-Y. Lin, ``{ROUGE}: A package for automatic evaluation of summaries,'' in
  \emph{Text Summarization Branches Out}, 2004, pp. 74--81.

\bibitem{37}
M.~Denkowski and A.~Lavie, ``Meteor universal: Language specific translation
  evaluation for any target language,'' in \emph{Proceedings of the Ninth
  Workshop on Statistical Machine Translation}, 2014, pp. 376--380.

\bibitem{38}
T.~Zhang, V.~Kishore, F.~Wu, K.~Q. Weinberger, and Y.~Artzi, ``{BERTScore}:
  Evaluating text generation with bert,'' in \emph{Proc. of ICLR}, 2020.

\bibitem{39}
S.~Watanabe, T.~Hori, S.~Karita, T.~Hayashi, J.~Nishitoba, Y.~Unno, N.~{Enrique
  Yalta Soplin}, J.~Heymann, M.~Wiesner, N.~Chen, A.~Renduchintala, and
  T.~Ochiai, ``{ESPnet}: End-to-end speech processing toolkit,'' in \emph{Proc.
  INTERSPEECH}, 2018, pp. 2207--2211.

\end{thebibliography}

\end{document}